\documentclass{article} 
\usepackage{iclr2027_conference,times}

\usepackage{amsmath,amssymb}

\usepackage{graphicx}
\usepackage{booktabs}
\usepackage{microtype}

\usepackage{float}
\usepackage{xurl}
\usepackage{hyperref}
\usepackage{algorithm}
\usepackage{algpseudocode}

\usepackage{array}
\usepackage{tabularx}
\usepackage{placeins}
\usepackage{enumitem}
\usepackage{colortbl}

\hypersetup{hidelinks}

\newcommand{\showcase}[2]{%
    \begin{figure}[ht]
        \centering
        \includegraphics[width=\linewidth]{figures/showcase_cls#1.png}
        \caption{
            \textbf{Additional uncurated samples.}
            Class #1: #2.
        }
        \label{fig:app-demo-#1}
    \end{figure}
}

\title{An End-to-End Latent-Rollout Approach for Pushing Few-Step ImageNet-$256$ Generation to FID $1.11$ without Fr{\'e}chet Losses}

\author{%
  Xiaoran Xu \\
  ByteDance
\And
  Yujing Wang \\
  ByteDance
}

\iclrfinalcopy 
\begin{document}

\maketitle
\lhead{Preprint}

\begin{abstract}
Iterative generation poses a joint optimization problem across steps, as intermediate predictions shape subsequent computations and ultimately determine the final output distribution.
Few-step generators distilled from pretrained diffusion and flow-matching models make such optimization computationally practical end to end.
We build on this opportunity with a distill-then-refine approach that uses teacher imitation to establish a strong initialization for a few-step rollout in latent space, then shifts to end-to-end refinement of the complete latent rollout against real data.
We introduce FiST (Flow-in-Stage Transformer), an architecture that composes learned latent-state transitions in a few stages using a shared Transformer, with optional cross-stage hidden communication.
Distillation applies teacher-forced regression to selected states along teacher trajectories; refinement replaces this supervision with adversarial and auxiliary classification objectives on the final latent output.
A trainable discriminator module operates on semantically rich features extracted from clean real and generated latents by a frozen SiT backbone pretrained with REPA.
All training takes place in latent space, without image decoding.
During refinement, FiST consumes its own intermediate predictions, and endpoint gradients pass through every generation stage.
For class-conditional generation on ImageNet at $256\times256$, our approach achieves FID 1.11 (IS 282) with three stages and FID 1.15 (IS 280) with two.
These results demonstrate competitive few-step generation through learned distribution-level supervision, without explicit Fr{\'e}chet-distance minimization.
Ablations characterize how distillation, pretrained checkpoint choices, refinement supervision, and cross-stage hidden communication affect generation quality.
\end{abstract}

\section{Introduction}

Diffusion and flow-matching models achieve high-quality image synthesis through iterative sampling, but repeated network evaluations make generation computationally expensive \citep{ho2020denoising,lipman2023flow}.
Distillation accelerates sampling by transferring the behavior of these pretrained models into few-step generators \citep{salimans2022progressive}.
Beyond sampling acceleration, distillation can provide a strong initialization that facilitates stable distribution-level optimization against real data through an adversarial objective \citep{lin2025apt}.
Treating the complete few-step rollout as a differentiable generator allows its intermediate computations to be optimized jointly through supervision on the final output.
Our central question is \emph{whether end-to-end refinement of a distilled few-step rollout against real data can close the gap to many-step generation}.

We introduce \textbf{FiST (Flow-in-Stage Transformer)}, an architecture that factorizes the noise-to-data mapping into a composition of learned latent-state transitions across a small number of stages. 
A shared Transformer implements these transitions, mapping the current latent state to the next one at each stage; the final stage produces the generator’s latent output.
The basic rollout passes latent states between stages, while an augmented rollout additionally carries hidden features through a cross-stage hidden router.
We use a \emph{distill-then-refine} procedure to train both rollout forms entirely in latent space, separating distillation-based initialization from subsequent end-to-end refinement.
Distillation applies teacher-forced regression to selected states along teacher trajectories, including intermediate states and the final latent output.

The refinement phase replaces the regression supervision with adversarial and auxiliary classification objectives driven by real data.
We feed latents of real images and the generator's latent outputs into a frozen SiT backbone pretrained with REPA \citep{ma2024sit,yu2025repa}.
This backbone provides a semantically rich feature space shared by real and generated samples, while a trainable discriminator module uses two output heads to assess sample realism and class consistency, respectively.
The objectives therefore supervise the final latent output, leaving intermediate stage states free to adapt during refinement.
The discriminator processes clean latents directly, avoiding image decoding during optimization.
By learning to distinguish real from generated samples, the discriminator provides adaptive supervision for distribution-level refinement without explicit Fr{\'e}chet-distance minimization or dependence on a specific evaluation metric.

During refinement, FiST is executed from noise using its own intermediate predictions as inputs to subsequent stages, matching its inference computation.
For each generator update, we retain the full gradient path from the endpoint objectives through the discriminator and every generation stage, including the hidden router in the augmented rollout.
The discriminator's frozen feature backbone is differentiated with respect to the generated latent.
The shared Transformer therefore receives gradient contributions from its use at every stage, each accounting for how that stage's prediction affects the final output through the remaining stages.

We evaluate our approach on ImageNet $256\times256$ for class-conditional image generation.
FiST achieves \textbf{FID 1.11} (IS 282) with three stages and \textbf{FID 1.15} (IS 280) with two.
Both achieve lower reported FID than two-step FACM (1.32) and one-step W-Flow (1.29) \citep{peng2026facm,han2026wflow}, and approach the reported FID of leading many-step models such as GenFirst (0.97) \citep{zheng2026genfirst} and SFD (1.04) \citep{pan2025sfd}.
Ablations examine distillation duration, teacher classifier-free guidance, auxiliary classification, cross-stage hidden communication, pretrained checkpoints, and stage count.
Training curves of FID and IS further characterize the refinement dynamics.
These studies show that longer distillation and REPA-based discriminator features improve FID significantly after refinement, with additional gains from auxiliary classification and cross-stage hidden communication.
Teacher guidance also affects the FID--IS tradeoff.

Our contributions are summarized as follows:
\begin{itemize}[leftmargin=*]
    \item We introduce FiST, a Transformer architecture for staged latent rollouts that realizes the noise-to-data mapping through a short sequence of learned latent-state transitions, with a basic rollout and an augmented variant supporting cross-stage hidden communication.

    \item We develop an effective distill-then-refine procedure that initializes FiST by regressing to selected teacher-trajectory states under teacher forcing, then refines the complete latent rollout end to end using real-data adversarial and auxiliary classification supervision in a frozen REPA feature space.

    \item We evaluate our end-to-end latent-rollout approach for class-conditional image generation on ImageNet at $256\times256$, demonstrating competitive two- and three-stage performance and characterizing the effects of architecture, distillation, pretrained checkpoints, and refinement supervision.
\end{itemize}

\section{Related Work}
\label{sec:related_work}

Distillation trains few-step generators using supervision from pretrained diffusion and flow-matching models.
Progressive distillation repeatedly halves the number of steps in a deterministic diffusion sampler \citep{salimans2022progressive}.
Consistency models support one- and few-step generation and can be trained either by distillation from pretrained models or directly from data \citep{song2023consistency}.
Adversarial objectives further connect fast generation with distribution-level supervision.
ADD combines diffusion-teacher supervision with an adversarial loss for few-step generation \citep{sauer2024add}.
CTM combines a consistency distillation loss with denoising score matching and adversarial losses \citep{kim2024ctm}.
DMD introduces score-based distribution matching \citep{yin2024dmd}, and DMD2 augments it with real-data adversarial training \citep{yin2024dmd2}.
APT separates consistency-based initialization from subsequent adversarial post-training of a one-step generator \citep{lin2025apt}, while D2O directly fine-tunes a diffusion-pretrained generator with an adversarial objective \citep{zheng2025d2o}.
We train FiST with a distill-then-refine procedure: teacher-forced regression to selected teacher-trajectory states initializes the generator, and subsequent end-to-end refinement replaces regression with real-data adversarial and auxiliary classification objectives.

Pretrained representations provide a basis for distribution-level supervision.
Projected GANs train discriminator heads on fixed image features \citep{sauer2021projected}.
LADD performs adversarial distillation in latent space, using frozen diffusion-teacher features extracted from re-noised generated latents \citep{sauer2024ladd}.
Representation alignment offers an externally guided way to shape the features used in generative modeling: REPA aligns diffusion Transformer representations with pretrained visual features \citep{yu2025repa}, and GAT aligns a trainable Transformer discriminator with frozen DINOv2 features \citep{hyun2026gat}.
W-Flow trains one-step generators through Sinkhorn-derived distribution updates in pretrained feature spaces \citep{han2026wflow}.
FD-loss directly minimizes Fr{\'e}chet discrepancies in pretrained feature spaces \citep{yang2026fdloss}, while AdvFD augments fixed-representation matching with an adversarially learned representation \citep{gao2026advfd}.
Our refinement procedure uses a frozen SiT backbone pretrained with REPA as a latent feature extractor, with trainable adversarial and auxiliary classification heads but no Fr{\'e}chet-distance losses.
The backbone processes clean real and generated latents directly, avoiding image decoding.

Optimizing a multi-step generator also depends on which parts of the sampling process remain in the gradient graph.
DRaFT differentiates a reward through an entire diffusion sampling chain, while DRaFT-K restricts backpropagation to the last few denoising steps \citep{clark2024draft}.
In few-step distillation, DMD2 trains its multi-step generator on its own intermediate outputs, simulating inference \citep{yin2024dmd2}, and TDM matches teacher and student trajectory distributions while restricting backpropagation to one ODE step \citep{luo2025tdm}.
During refinement, we optimize FiST across its complete staged rollout using endpoint adversarial and auxiliary classification losses.
Gradients pass through the discriminator's frozen feature backbone and every generation stage of the generator, making earlier stages responsive to their effect on the final output.

\section{Approach}
\label{sec:approach}

\subsection{FiST Architecture}
\label{sec:fist_architecture}

Let $z=E(x)$ be the latent encoding of an image $x$ produced by a pretrained VAE encoder $E$, and let $c$ denote its class label.
FiST factorizes the noise-to-data mapping into a composition of learned latent-state transitions implemented by a shared Transformer $F_{\theta}$, with each forward pass constituting a \emph{generation stage}.
For Gaussian noise $\epsilon\sim\mathcal{N}(0,I)$ and a fixed stage count $K$, the basic rollout is:
\begin{equation}
    z_0=\epsilon,\qquad
    z_s=F_{\theta}(z_{s-1};e_s,c),\qquad s=1,\ldots,K,
    \label{eq:fist_basic_rollout}
\end{equation}
where $e_s$ is a learned embedding for stage $s$ and $z_K$ is the final latent output.
We construct $F_{\theta}$ from a pretrained SiT \citep{ma2024sit} by replacing its timestep embedder with a stage embedding table.
We also reinitialize its final prediction layer, as FiST predicts the next-stage state rather than velocity.
By default, FiST reuses backbone weights from a SiT checkpoint pretrained with REPA \citep{yu2025repa}.

The basic rollout passes latent states between stages.
An augmented rollout additionally retains layer-wise hidden features for subsequent stages through a cross-stage hidden router.
Let $H_s$ collect the hidden features retained from stage $s$, and let $\mathcal{H}_{<s}=\{H_j\}_{j<s}$ be the memory available to that stage.
With router parameters $\phi$ and $\Theta=(\theta,\phi)$, the augmented computation is
\begin{equation}
    (z_s,H_s)
    =F_{\Theta}(z_{s-1},\mathcal{H}_{<s};e_s,c),
    \qquad \mathcal{H}_{<1}=\varnothing.
    \label{eq:fist_augmented_rollout}
\end{equation}
Executing either rollout for $K$ stages from $z_0=\epsilon$ defines the complete differentiable generator $G_{\Theta}(\epsilon,c)=z_K$, where $\Theta=\theta$ for the basic rollout and $\Theta=(\theta,\phi)$ for the augmented rollout.
We develop a lightweight implementation of the router with negligible overhead in both parameter count and computation.
At each Transformer layer, attention operates along the stage axis at corresponding patch positions, with the current stage providing the query and the current and earlier stages providing the keys.
Because attention spans only a few stages, we use a single head with low-dimensional queries and keys.
Earlier-stage hidden features serve as values after learned channel-wise gating, avoiding a costly value projection.
The resulting message is added residually to the current hidden features.
Appendix~\ref{app:fist_architecture} provides further details on parameter initialization, routing, and overhead.

\subsection{Distillation-Based Initialization}
\label{sec:fist_distillation}

For each noise--label pair $(\epsilon,c)$, a frozen teacher generates a latent trajectory from noise to a latent endpoint.
We simply divide the sampling time interval into $K$ equal parts and select the corresponding trajectory states $\{\bar z_s\}_{s=0}^{K}$, with $\bar z_0=\epsilon$ and $\bar z_K$ the final latent.
FiST is then trained through teacher-forced regression to these selected states:
\begin{equation}
    \begin{aligned}
        (\hat z_s,H_s)
        &=F_{\Theta}(\bar z_{s-1},\mathcal{H}_{<s};e_s,c),
        \qquad s=1,\ldots,K,\\
        \mathcal{L}_{\mathrm{dist}}(\Theta)
        &=\mathbb{E}_{\epsilon,c}\!\left[
        \sum_{s=1}^{K}\operatorname{MSE}(\hat z_s,\bar z_s)
        \right].
    \end{aligned}
    \label{eq:fist_distillation}
\end{equation}
The loss supervises FiST's predictions of the selected intermediate states and the final latent output.
Each stage receives the preceding teacher state $\bar z_{s-1}$ as its latent input; its prediction $\hat z_s$ is regressed to $\bar z_s$ and does not become the next stage's input.
In the augmented rollout, $\mathcal{H}_{<s}$ contains FiST-computed hidden features from earlier teacher-forced stages, preserving differentiable cross-stage hidden communication; the basic rollout omits the hidden-memory inputs and outputs.
End-to-end refinement of the complete latent rollout starts from the distilled FiST checkpoint.
Appendix~\ref{app:fist_distillation} provides implementation details and the distillation algorithm.

\subsection{End-to-End Refinement}
\label{sec:fist_refinement}

Starting from the distilled checkpoint, we execute FiST from noise using its own intermediate predictions as inputs to subsequent stages.
The rollout matches the inference computation and produces the final latents $z^{-}=G_{\Theta}(\epsilon,c)$, while real image latents $z^{+}=E(x)$ serve as samples from the target distribution.
Refinement replaces teacher-forced regression with adversarial and auxiliary classification objectives on this latent endpoint.
The teacher and its selected trajectory states are no longer used.

\paragraph{Latent-space discriminator.}
We extract features from both real image latents $z^{+}$ and generated endpoints $z^{-}$ using a frozen SiT backbone pretrained with REPA \citep{yu2025repa}.
The backbone processes clean latents at $t=0$, providing semantically rich multi-layer hidden features.
Let $\Phi(z,\tilde c)$ collect the feature maps from selected layers, where $\tilde c$ equals $c$ or the unconditional label according to the conditioning dropout used in backbone pretraining.
This exposes the discriminator's trainable module to both conditional and unconditional features during training, with both objectives using the same features from a single backbone pass.
The module $\mathcal{D}_{\omega}$ consists of a convolutional aggregation network followed by two output heads that produce an adversarial logit $d_{\omega}$ and a vector of class logits $q_{\omega}$:
\begin{equation}
    \bigl(d_{\omega}(z,\tilde c),q_{\omega}(z,\tilde c)\bigr)
    =\mathcal{D}_{\omega}\bigl(\Phi(z,\tilde c)\bigr).
    \label{eq:fist_discriminator_outputs}
\end{equation}
Here $\omega$ denotes all trainable parameters, while the backbone parameters remain frozen.
The original class label $c$ remains the classification target.
Appendix~\ref{app:fist_refinement} details the selected feature layers, conditioning dropout, and trainable module architecture.

\paragraph{Refinement objectives.}
We use a logistic adversarial objective with a non-saturating generator loss \citep{goodfellow2014gan}, together with auxiliary classification supervision \citep{odena2017acgan}.
Write $\operatorname{sp}(a)=\log(1+\exp(a))$ for softplus, $\operatorname{sg}$ for stop-gradient, and $\operatorname{CE}$ for cross-entropy over class logits.
Expectations include the randomness of conditioning dropout.
For the discriminator update, we minimize
\begin{equation}
    \mathcal{L}_{\mathrm{D}}(\omega;\Theta)
    =\mathbb{E}_{z^{+},c,\tilde c}\!\left[
    \operatorname{sp}\bigl(-d_{\omega}(z^{+},\tilde c)\bigr)
    +\lambda_{\mathrm{cls}}^{\mathrm{D}}
    \operatorname{CE}\bigl(q_{\omega}(z^{+},\tilde c),c\bigr)
    \right]
    +\mathbb{E}_{\epsilon,c,\tilde c}\!\left[
    \operatorname{sp}\bigl(d_{\omega}(\operatorname{sg}[z^{-}],\tilde c)\bigr)
    \right].
    \label{eq:fist_discriminator_loss}
\end{equation}
The adversarial term trains the discriminator to distinguish real from generated samples, while the classification term is applied only to real samples.
For the generator update, both objectives act on the final latent output:
\begin{equation}
    \mathcal{L}_{\mathrm{G}}(\Theta;\omega)
    =\mathbb{E}_{\epsilon,c,\tilde c}\!\left[
    \operatorname{sp}\bigl(-d_{\omega}(G_{\Theta}(\epsilon,c),\tilde c)\bigr)
    +\lambda_{\mathrm{cls}}^{\mathrm{G}}
    \operatorname{CE}\bigl(q_{\omega}(G_{\Theta}(\epsilon,c),\tilde c),c\bigr)
    \right].
    \label{eq:fist_generator_loss}
\end{equation}
The two loss terms encourage sample realism and consistency with the requested class, respectively.
We use $\lambda_{\mathrm{cls}}^{\mathrm{D}}=0.7$ and $\lambda_{\mathrm{cls}}^{\mathrm{G}}=0.5$; Appendix~\ref{app:fist_refinement} explains how the loss weights are set to balance initial gradient magnitudes.
These losses require neither paired real targets for generated samples nor regression to teacher outputs.
In other words, given a noise--label pair $(\epsilon,c)$, the generated endpoint is not required to reproduce a particular real image or teacher prediction.
Instead, the two objectives supervise the output distribution through a discriminator trained on real and generated samples.

\paragraph{End-to-end updates.}
Each refinement iteration takes one discriminator update followed by one generator update.
During the generator update, $\omega$ is held fixed and we retain the full gradient path through $\mathcal{D}_{\omega}$, the frozen feature backbone $\Phi$, and every generation stage.
Let $\ell_{\mathrm{G}}(z,c,\tilde c;\omega)$ denote the per-sample loss inside the expectation in Eq.~\eqref{eq:fist_generator_loss}, evaluated at generated latent $z$.
The generator gradient is
\begin{equation}
    \nabla_{\Theta}\mathcal{L}_{\mathrm{G}}
    =\mathbb{E}_{\epsilon,c,\tilde c}\!\left[
    \left(\frac{\partial G_{\Theta}(\epsilon,c)}{\partial\Theta}\right)^{\!\mathsf T}
    \nabla_{z}\ell_{\mathrm{G}}(z,c,\tilde c;\omega)
    \big|_{z=G_{\Theta}(\epsilon,c)}
    \right].
    \label{eq:fist_endpoint_gradient}
\end{equation}
Computing the endpoint gradient requires differentiating $\Phi$ with respect to its latent input while keeping the backbone parameters fixed.
The rollout Jacobian accounts for all intermediate latent-state dependencies and, in the augmented rollout, cross-stage hidden-memory dependencies.

To make each stage's contribution to the shared-parameter gradient explicit, consider the basic rollout and write $F_{\theta}^{s}(z)=F_{\theta}(z;e_s,c)$.
Let $J_s=\partial F_{\theta}^{s}/\partial z_{s-1}$ be its input Jacobian and $A_s=\partial F_{\theta}^{s}/\partial\theta$ its parameter Jacobian with the stage input held fixed, both evaluated along the current rollout.
For one sampled endpoint loss, the state gradients $v_s$ and parameter gradient satisfy
\begin{equation}
    v_K=\nabla_{z_K}\ell_{\mathrm{G}},\qquad
    v_{s-1}=J_s^{\mathsf T}v_s,\quad s=K,\ldots,1,\qquad
    \nabla_{\theta}\ell_{\mathrm{G}}=\sum_{s=1}^{K}A_s^{\mathsf T}v_s.
    \label{eq:fist_basic_shared_gradient}
\end{equation}
The shared Transformer therefore receives a contribution from every stage invocation, with each $v_s$ accounting for how that stage's prediction influences the final output through the remaining stages.
Later stages take the generator's own predictions as inputs, and end-to-end updates jointly optimize the computations that produce and process these predictions.
For the augmented rollout, the same recursion applies after including the hidden memory in the stage state and using the full parameter set $\Theta$.
This additionally accounts for the effect of earlier hidden features on all later stages that access them through the router.

With supervision applied only at the endpoint, intermediate states can depart from teacher states as the complete latent rollout is optimized for final sample quality.
All refinement objectives are evaluated in latent space, without image decoding or Fr{\'e}chet-distance losses.
Appendix~\ref{app:fist_refinement} provides the refinement algorithm and details gradient propagation through both rollout forms.
\section{Experiments}

\subsection{Experimental Setup}
\label{sec:experimental_setup}

We study class-conditional image generation on ImageNet at $256\times256$ resolution.
Images are center-cropped and resized, then encoded using the pretrained SD-VAE \citep{rombach2022ldm}; no additional data augmentation is applied.
Our default configuration uses the officially released SiT-XL/2 checkpoint trained with REPA \citep{yu2025repa} as the teacher and as the source of pretrained weights for FiST and the discriminator backbone.

FiST is trained for 40K distillation steps followed by 20K refinement iterations, with a batch size of 1024 throughout.
Distillation uses a learning rate of $10^{-4}$.
Refinement alternates one discriminator update with one generator update, using respective learning rates of $4\times10^{-5}$ and $10^{-5}$ following the two-time-scale update rule (TTUR) \citep{heusel2017ttur}.
Both training phases use AdamW and cosine learning-rate decay with a 1K-step warmup.
Detailed hyperparameters are provided in Appendix~\ref{app:experimental_setup}.

Evaluation follows the ADM protocol \citep{dhariwal2021diffusion} using 50K generated images with randomly sampled class labels.
We report FID, IS, sFID, precision, and recall at the lowest-FID
FiST checkpoint within 20K refinement iterations.
FiST uses neither inference-time classifier-free guidance nor an exponential moving average of generator parameters.
We report FiST's stage count, parameter count, and inference FLOPs in Table~\ref{tab:fist_compute}.

\subsection{Comparison with Existing Methods}
\label{sec:comparison}

\paragraph{Model size and inference cost.}
Table~\ref{tab:fist_compute} summarizes FiST's model size and inference cost.
Sharing Transformer weights across stages allows longer rollouts with little change in parameter count, while computation grows approximately linearly with stage count.
For the three-stage configuration, cross-stage hidden communication adds approximately 4M parameters and 3.6 GFLOPs per image, increasing model size by about $0.6\%$ and computation by about $1.0\%$.
An augmented FiST stage therefore costs approximately one SiT-XL/2 evaluation, allowing the stage counts in Table~\ref{tab:imagenet256_comparison} to be interpreted in terms of SiT-equivalent compute.

\begin{table}[t]
\centering
\caption{\textbf{Model size and inference cost of FiST.}
Augmented FiST includes the cross-stage hidden router.
Parameter counts and FLOPs are approximate.
Inference costs exclude VAE decoding.
For reference, SiT-XL/2 has approximately 675M parameters and costs 118.7 GFLOPs per evaluation.
SiT-equivalent compute per stage is the average per-stage cost relative to one such evaluation.}
\label{tab:fist_compute}
\begingroup
\small
\setlength{\tabcolsep}{3pt}
\renewcommand{\arraystretch}{1.08}
\begin{tabular*}{\linewidth}{@{\extracolsep{\fill}}l c r r r r@{}}
\toprule
Type & Stages & \shortstack[r]{Parameters\\(M)} & \shortstack[r]{GFLOPs\\per image} & \shortstack[r]{Avg. GFLOPs\\per stage} & \shortstack[r]{SiT-equivalent\\compute per stage} \\
\midrule
Basic & 3 & 673 & 356.1 & 118.7 & $1.00\times$ \\
Augmented & 2 & 677 & 239.8 & 119.9 & $1.01\times$ \\
Augmented & 3 & 677 & 359.7 & 119.9 & $1.01\times$ \\
\bottomrule
\end{tabular*}
\endgroup
\end{table}

\begin{table}[t]
\centering
\caption{\textbf{System-level comparison on class-conditional ImageNet $256\times256$.}
Baseline results follow the cited works and official releases.
NFE denotes nominal network evaluations; Stg.\ denotes FiST stages (see Table~\ref{tab:fist_compute} for costs).
SD: SD-VAE; E2E: VAE trained jointly with the generator; Sem: SD-VAE + SemVAE; RAE: representation autoencoder.
SIM: SigLIP + Inception + MAE.}
\label{tab:imagenet256_comparison}
\begingroup
\small
\setlength{\tabcolsep}{2.6pt}
\renewcommand{\arraystretch}{1.06}
\begin{tabularx}{\linewidth}{@{}>{\raggedright\arraybackslash}X c c r r r r r@{}}
\toprule
Method & Space & NFE / Stg. & FID$\downarrow$ & IS$\uparrow$ & sFID$\downarrow$ & Prec.$\uparrow$ & Rec.$\uparrow$ \\
\midrule
\multicolumn{8}{@{}l}{\textit{Many-step generation without Fr{\'e}chet losses}} \\
DiT-XL/2 \citep{peebles2023dit} & SD & $250\times2$ & 2.27 & 278.2 & 4.60 & 0.83 & 0.57 \\
SiT-XL/2 \citep{ma2024sit} & SD & $250\times2$ & 2.06 & 277.5 & 4.49 & 0.83 & 0.59 \\
SiT-XL/2 + REPA \citep{yu2025repa} & SD & $250\times2$ & 1.42 & 305.7 & 4.70 & 0.80 & 0.65 \\
SiT-XL/1 + REPA-E \citep{leng2025repae} & E2E & $250\times2$ & 1.12 & 302.9 & 4.09 & 0.79 & 0.66 \\
RAE + DiT$^{\mathrm{DH}}$-XL \citep{zheng2025rae} & RAE & $50\times2$ & 1.13 & 262.6 & -- & 0.78 & 0.67 \\
UCGM-S (REPA-E) \citep{sun2025unified} & E2E & $40\times2$ & 1.06 & -- & -- & -- & -- \\
SFD-XXL (Euler) \citep{pan2025sfd} & Sem & $200\times2$ & 1.04 & -- & -- & -- & -- \\
GenFirst (EiT) \citep{zheng2026genfirst} & E2E & $250\times2$ & 0.97 & 283.4 & -- & 0.79 & 0.67 \\
\midrule
\multicolumn{8}{@{}l}{\textit{One- and few-step generation without Fr{\'e}chet losses}} \\
MeanFlow-XL/2 \citep{geng2025meanflow,geng2025improved} & SD & 1 & 3.43 & 247.5 & -- & -- & -- \\
iMF-XL/2 \citep{geng2025improved} & SD & 1 & 1.72 & 282.0 & -- & -- & -- \\
iMF-XL/2 \citep{geng2025improved} & SD & 2 & 1.54 & -- & -- & -- & -- \\
AFM-XL/2 \citep{lin2025afm} & SD & 2 & 2.11 & 273.8 & 4.33 & 0.82 & 0.55 \\
AFM-XL/2, 112 layers \citep{lin2025afm} & SD & 1 & 1.94 & 292.2 & 4.54 & 0.79 & 0.56 \\
Drifting-L/2 \citep{deng2026drifting} & SD & 1 & 1.54 & 258.9 & -- & -- & -- \\
W-Flow-XL/2 \citep{han2026wflow} & SD & 1 & 1.29 & 265.4 & -- & -- & -- \\
UCGM-T (REPA-E) \citep{sun2025unified} & E2E & 2 & 1.39 & -- & -- & -- & -- \\
FACM-XL \citep{peng2026facm} & VA & 2 & 1.32 & 292.0 & -- & -- & -- \\
FreeFlow-XL/2 \citep{tong2026freeflow} & SD & 1 & 1.45 & -- & -- & -- & -- \\
\midrule
\multicolumn{8}{@{}l}{\textit{One-step generation with Fr{\'e}chet losses}} \\
iMF-XL/2 + FD (Inception) \citep{yang2026fdloss} & SD & 1 & 0.72 & 295.0 & -- & 0.76 & 0.68 \\
iMF-XL/2 + FD (SIM) \citep{yang2026fdloss} & SD & 1 & 0.76 & 301.3 & -- & 0.77 & 0.67 \\
JiT-H + AdvFD \citep{gao2026advfd} & Pixel & 1 & 0.72 & -- & -- & -- & -- \\
\midrule
\multicolumn{8}{@{}l}{\textit{FiST: distill-then-refine without Fr{\'e}chet losses}} \\
\rowcolor[gray]{0.94}
FiST (basic) & SD & 3 & 1.16 & 282.6 & 4.19 & 0.76 & 0.68 \\
\rowcolor[gray]{0.94}
FiST (augmented) & SD & 2 & 1.15 & 280.1 & 4.18 & 0.76 & 0.69 \\
\rowcolor[gray]{0.94}
FiST (augmented) & SD & 3 & \textbf{1.11} & 282.0 & 4.17 & 0.76 & 0.69 \\
\bottomrule
\end{tabularx}
\endgroup
\end{table}

\paragraph{Quantitative comparison.}
Table~\ref{tab:imagenet256_comparison} compares basic and augmented FiST with existing image generators.
The three-stage basic rollout achieves FID 1.16 with IS 282.6.
The augmented rollout achieves FID 1.15 with IS 280.1 using two stages and FID 1.11 with IS 282.0 using three.
All three configurations obtain lower reported FID than the listed one- and few-step methods trained without Fr{\'e}chet losses, including W-Flow, UCGM-T and FACM.
These results are also comparable to those of many-step models such as REPA-E (FID 1.12) and RAE (FID 1.13), which use $250\times2$ and $50\times2$ network evaluations, respectively.
Several many-step models and methods trained with Fr{\'e}chet losses report lower FID.
Nevertheless, FiST demonstrates that competitive FID can be achieved with a short latent rollout through adversarial distribution learning and auxiliary classification, without explicit Fr{\'e}chet-distance minimization.
Our discriminator adapts its supervision to discrepancies between real and generated samples, offering a flexible route to distribution-level optimization that does not depend on a differentiable formulation of the evaluation metric.

The additional metrics provide a broader view of generation quality.
FiST's sFID scores of 4.17--4.19 are lower than those of the listed DiT, SiT, REPA, and AFM configurations, but slightly higher than REPA-E's 4.09.
All three FiST configurations achieve precision 0.76, with recall 0.68 for the basic rollout and 0.69 for both augmented configurations.
Although adversarial training can suffer from mode collapse, the augmented FiST configurations achieve the highest reported recall among the listed methods.
This suggests that distillation-based initialization followed by end-to-end adversarial refinement can attain low FID together with broad distribution coverage.
The lower precision relative to several baselines nevertheless reflects a remaining tradeoff between coverage and fidelity.

\begin{figure}[t]
    \centering
    \includegraphics[width=\linewidth]{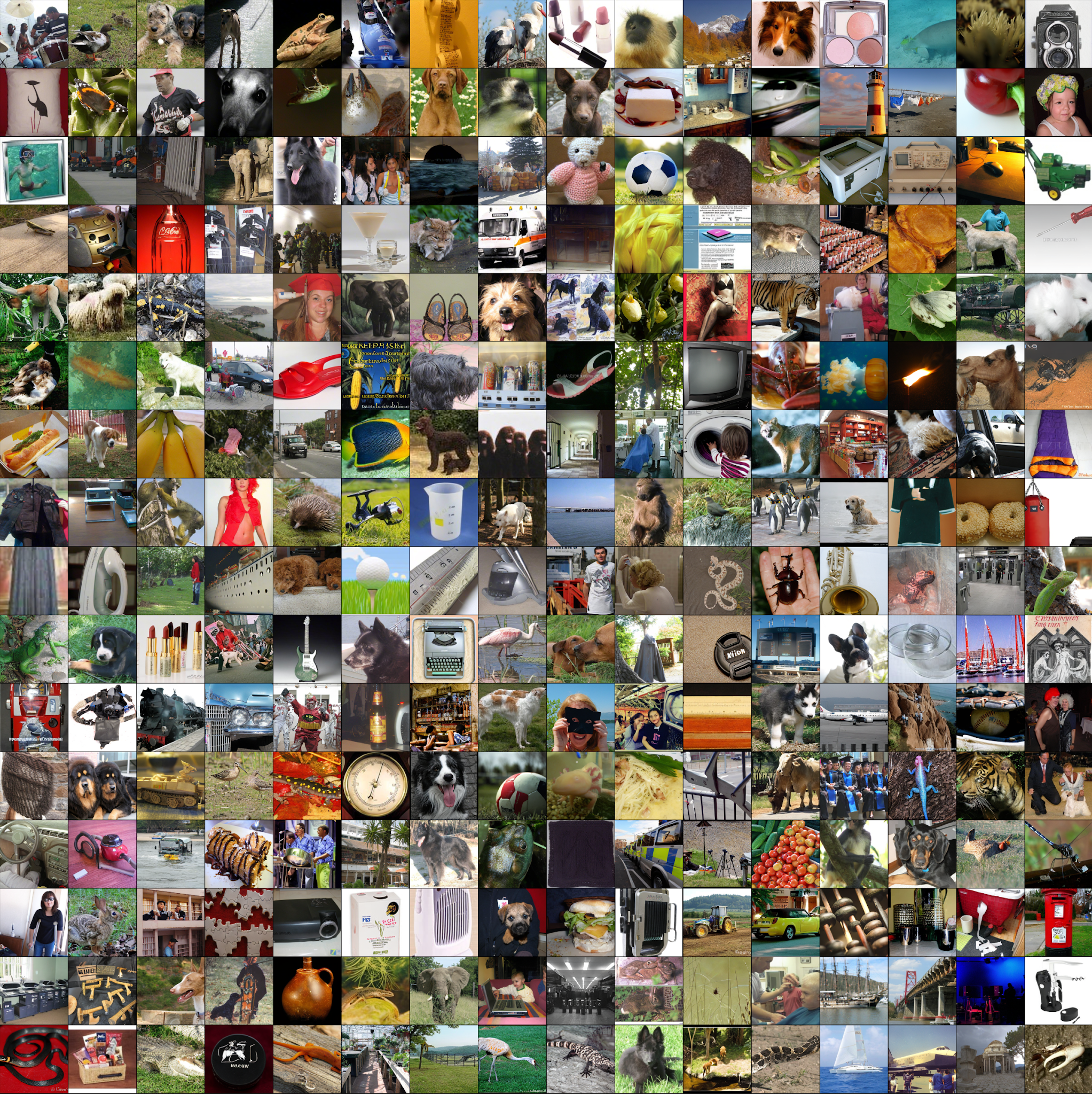}
    \caption{\textbf{Uncurated samples from three-stage augmented FiST on ImageNet $256\times256$.}
    Samples are generated from randomly drawn Gaussian noise and class labels.}
    \label{fig:fist_samples}
\end{figure}

\paragraph{Qualitative results.}
Figure~\ref{fig:fist_samples} shows uncurated samples spanning animals, everyday objects, vehicles, food, and scenes, with rich texture detail and a natural photographic appearance.
Most samples depict recognizable objects with coherent structures and plausible scene layouts.
This visual quality is obtained from REPA-pretrained weights with 40K distillation steps and 20K refinement iterations, using ImageNet as the sole image dataset for refinement.

\subsection{Distillation and Refinement Dynamics}
\label{sec:fist_refinement_dynamics}

\begin{table}[t]
\centering
\caption{\textbf{Effects of distillation duration and teacher guidance.}
Results use three-stage augmented FiST.
Teacher guidance uses REPA's best-FID and qualitative-sampling scales, 1.8 and 4.0, with guidance intervals $[0,0.7]$ and $[0,1]$, respectively \citep{yu2025repa}.
Arrows indicate before $\rightarrow$ after refinement; metrics are measured at the lowest-FID checkpoint within 20K refinement iterations.}
\label{tab:fist_distillation_ablation}
\begingroup
\small
\setlength{\tabcolsep}{3pt}
\renewcommand{\arraystretch}{1.08}

\begin{minipage}[t]{0.46\linewidth}
\vspace{0pt}
\centering
\textit{Distillation duration (CFG 1.8, $[0,0.7]$)}
\par\smallskip
\begin{tabular*}{\linewidth}{@{\extracolsep{\fill}}l c c@{}}
\toprule
Steps & FID$\downarrow$ & IS$\uparrow$ \\
\midrule
4K  & $60.58 \rightarrow 1.69$ & $35.7 \rightarrow 239.7$ \\
10K & $8.76 \rightarrow 1.19$ & $172.1 \rightarrow 271.7$ \\
40K & $5.23 \rightarrow 1.11$ & $205.7 \rightarrow 282.0$ \\
\bottomrule
\end{tabular*}
\end{minipage}
\hfill
\begin{minipage}[t]{0.50\linewidth}
\vspace{0pt}
\centering
\textit{Teacher guidance (40K distillation)}
\par\smallskip
\begin{tabular*}{\linewidth}{@{\extracolsep{\fill}}l c c c@{}}
\toprule
CFG & Interval & FID$\downarrow$ & IS$\uparrow$ \\
\midrule
1.8 & $[0,0.7]$ & $5.23 \rightarrow 1.11$
& $205.7 \rightarrow 282.0$ \\
4.0 & $[0,1]$ & $10.41 \rightarrow 1.23$
& $448.8 \rightarrow 302.5$ \\
\bottomrule
\end{tabular*}
\end{minipage}

\endgroup
\end{table}

\begin{figure}[t]
    \centering
    \includegraphics[width=\linewidth]{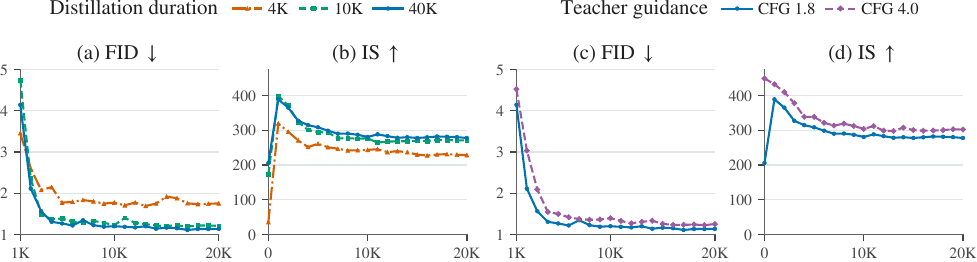}
    \caption{\textbf{Refinement dynamics under different distillation and teacher guidance settings.}
    Experimental settings match Table~\ref{tab:fist_distillation_ablation}; horizontal axes show refinement iterations.
    FID curves begin at 1K to prevent large pre-refinement values from obscuring later differences.}
    \label{fig:fist_refinement_dynamics}
\end{figure}

We study how distillation duration and teacher guidance affect refinement of three-stage augmented FiST.
Table~\ref{tab:fist_distillation_ablation} summarizes outcomes, while Figure~\ref{fig:fist_refinement_dynamics} tracks their evolution during refinement.

\paragraph{Distillation duration.}
We compare 4K, 10K, and 40K distillation steps under a common learning-rate schedule, each followed by the same 20K-iteration refinement procedure.
Refinement improves all three checkpoints, reducing FID from 60.58 to a best of 1.69 even after only 4K distillation steps.
Longer distillation lowers the best FID to 1.19 at 10K and 1.11 at 40K, with diminishing gains.
The FID gaps persist late in refinement, suggesting that the preceding distillation still affects quality under a fixed refinement budget.

\paragraph{Teacher guidance.}
With 40K distillation steps, we compare teacher CFG 1.8 over $[0,0.7]$ against CFG 4.0 over $[0,1]$.
Stronger guidance yields higher IS but worse FID before and after refinement: its best-FID checkpoint reaches FID 1.23 with IS 302.5, versus FID 1.11 and IS 282.0 for CFG 1.8.
Teacher guidance thus affects refinement even after teacher supervision ends.
For CFG 1.8, Figure~\ref{fig:fist_refinement_dynamics} shows an early IS peak followed by a decline while FID generally improves.
IS therefore peaks well before FID reaches its minimum.
The early IS rise may reflect auxiliary classification strengthening class-discriminative cues; see Section~\ref{sec:fist_component_ablations} for analysis.

\subsection{Component Ablations}
\label{sec:fist_component_ablations}

\begin{figure}[t]
    \centering
    \includegraphics[width=\linewidth]{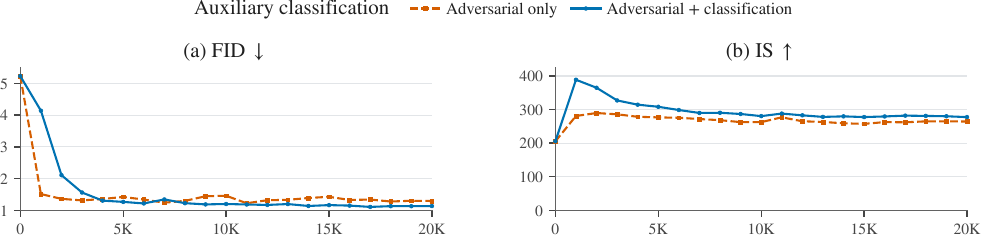}
    \caption{\textbf{Auxiliary classification during refinement.}
    Horizontal axes show refinement iterations; iteration 0 denotes the distilled checkpoint before refinement.}
    \label{fig:fist_aux_classification}
\end{figure}

\paragraph{Auxiliary classification.}
Adding auxiliary classification improves the best FID from 1.22 to 1.11 and the corresponding IS from 277.0 to 282.0 (Table~\ref{tab:fist_component_ablations}).
The training curves reveal different early behavior: adversarial-only refinement reduces FID faster, whereas classification produces a larger initial IS rise (Figure~\ref{fig:fist_aux_classification}).
Later in refinement, classification yields both lower FID and higher IS, indicating that its benefit extends beyond the transient IS increase.

\paragraph{Pretrained checkpoints.}
Table~\ref{tab:fist_component_ablations} compares vanilla SiT and REPA checkpoints.
With a vanilla SiT checkpoint used for both distillation (as the teacher and to initialize FiST) and the discriminator’s frozen backbone, subsequent refinement does not improve the distilled generator’s FID, which stays above 8.
Pairing a SiT checkpoint for distillation with a REPA checkpoint for the discriminator yields FID 1.21, while using REPA checkpoints for both yields 1.11.
These results highlight the importance of the discriminator's pretrained features, with a further gain from REPA-based distillation.

\paragraph{Cross-stage hidden communication.}
At three stages, the augmented rollout improves FID from 1.16 to 1.11 over the basic rollout (Table~\ref{tab:imagenet256_comparison}).
This modest gain requires approximately 1\% additional compute per stage (Table~\ref{tab:fist_compute}), making hidden communication an inexpensive architectural addition.

\paragraph{Stage count.}
We also experiment with one- and four-stage FiST, obtaining FID 1.47 and 1.12, respectively.
The higher FID with one stage supports the benefit of multiple learned transitions, while the fourth stage adds computation without improving on the three-stage result.
Together with the results above, these findings support two- and three-stage configurations as effective choices for balancing generation quality and inference cost.

\begin{table}[t]
\centering
\caption{\textbf{Effects of auxiliary classification and pretrained checkpoints.}
Results use three-stage augmented FiST.
Left: FID and IS at the lowest-FID checkpoint within 20K refinement iterations.
Right: FID with different pretrained weights for distillation (teacher and FiST initialization) and the discriminator's frozen backbone.
SiT and REPA denote vanilla and REPA-trained SiT-XL/2 checkpoints, respectively.
$\dagger$Refinement does not improve the distilled generator's FID.}
\label{tab:fist_component_ablations}
\begingroup
\small
\setlength{\tabcolsep}{3pt}
\renewcommand{\arraystretch}{1.08}

\begin{minipage}[t]{0.48\linewidth}
\centering
\textit{Auxiliary classification}\par\smallskip
\begin{tabular*}{\linewidth}{@{\extracolsep{\fill}}lrr@{}}
\toprule
Refinement objectives & FID$\downarrow$ & IS$\uparrow$ \\
\midrule
Adversarial only & 1.22 & 277.0 \\
Adversarial + classification & 1.11 & 282.0 \\
\bottomrule
\end{tabular*}
\end{minipage}
\hfill
\begin{minipage}[t]{0.48\linewidth}
\centering
\textit{Pretrained checkpoints}\par\smallskip
\begin{tabular*}{\linewidth}{@{\extracolsep{\fill}}llr@{}}
\toprule
Distillation & Discriminator & FID$\downarrow$ \\
\midrule
SiT & SiT & $>8^{\dagger}$ \\
SiT & REPA & 1.21 \\
REPA & REPA & 1.11 \\
\bottomrule
\end{tabular*}
\end{minipage}
\endgroup
\end{table}

\section{Conclusion}

We presented FiST and a distill-then-refine approach for optimizing few-step image generation end to end in latent space.
Teacher imitation establishes staged latent rollouts, after which adversarial and auxiliary classification objectives refine them against real data.
On class-conditional ImageNet $256\times256$, FiST achieves FID 1.15 with two stages and 1.11 with three, without image decoding during training or explicit Fr{\'e}chet-distance minimization.
Ablations highlight the importance of distillation and pretrained checkpoints, with further gains from auxiliary classification and cross-stage hidden communication.
These results support short differentiable rollouts initialized by distillation as a practical basis for end-to-end optimization through adaptive distribution-level supervision.

\clearpage
\bibliography{reference}
\bibliographystyle{iclr2027_conference}

\clearpage
\appendix
\section{FiST Architecture Details}
\label{app:fist_architecture}

\paragraph{Parameter initialization.}
FiST retains SiT's pretrained patch embedding, class-conditioning pathway, and Transformer blocks.
The timestep embedder is replaced by a learned stage embedding table $E_{\mathrm{stage}}\in\mathbb{R}^{K\times D}$, where $D$ is the Transformer hidden width; the selected embedding $e_s=E_{\mathrm{stage}}[s]$ enters the existing conditioning pathway.
The stage embeddings and final prediction layer are initialized from scratch, as are the router parameters in the augmented rollout.

\paragraph{Cross-stage hidden router.}
\label{app:fist_router}
The router is designed to add cross-stage information flow at negligible overhead.
Let $h_{s,\ell,n}\in\mathbb{R}^{D}$ denote the hidden feature at stage $s$, layer $\ell$, and patch position $n$.
For $s\geq2$, the router applies single-head causal attention at each Transformer layer, operating across stages at corresponding patch positions.
Spatial interactions between patch positions remain part of the Transformer's within-stage computation.
The query and key projections are affine, with output width $r=64$; no normalization is applied before or after these projections.
Their weights $W_{\ell}^{Q},W_{\ell}^{K}\in\mathbb{R}^{r\times D}$ and biases $b_{\ell}^{Q},b_{\ell}^{K}\in\mathbb{R}^{r}$ are shared across stages and patch positions.
The projected queries and keys determine scaled dot-product attention weights over the current and earlier stages:
\begin{equation}
\begin{aligned}
    q_{s,\ell,n}
    &=W_{\ell}^{Q}h_{s,\ell,n}+b_{\ell}^{Q},
    \qquad
    k_{j,\ell,n}=W_{\ell}^{K}h_{j,\ell,n}+b_{\ell}^{K},\\
    \alpha_{s\leftarrow j,\ell,n}
    &=
    \frac{\exp\!\left(q_{s,\ell,n}^{\mathsf T}k_{j,\ell,n}/\sqrt{r}\right)}
    {\sum_{i=1}^{s}\exp\!\left(q_{s,\ell,n}^{\mathsf T}k_{i,\ell,n}/\sqrt{r}\right)},
    \quad j=1,\ldots,s.
\end{aligned}
\label{eq:fist_router_attention}
\end{equation}
A learned channel-wise gate $g_{s\leftarrow j,\ell}\in\mathbb{R}^{D}$ is associated with each directed stage pair and layer, and shared across patch positions.
The router aggregates gated hidden features from earlier stages and adds the resulting message residually:
\begin{equation}
\begin{aligned}
    m_{s,\ell,n}
    &=\sum_{j=1}^{s-1}
    \alpha_{s\leftarrow j,\ell,n}
    \bigl(g_{s\leftarrow j,\ell}\odot h_{j,\ell,n}\bigr),\\
    \widetilde h_{s,\ell,n}
    &=h_{s,\ell,n}+m_{s,\ell,n},
\end{aligned}
\label{eq:fist_router_message}
\end{equation}
where $\odot$ denotes element-wise multiplication.
The routed features $\widetilde h_{s,\ell,n}$ are passed to Transformer block $\ell$,
while $H_s$ retains the features $h_{s,\ell,n}$ before routing for use by later stages.
The current-stage key participates in the softmax, but only earlier stages contribute values; the remaining weights are not renormalized.
No value or output projection is used, and the first stage receives no message.

\paragraph{Parameter and computational overhead.}
\label{app:fist_overhead}
For $L$ Transformer layers and $K\geq2$ stages, the router parameter count is
\begin{equation}
    P_{\mathrm{router}}
    =L\left[2r(D+1)+D\frac{K(K-1)}{2}\right],
    \label{eq:fist_router_parameters}
\end{equation}
where the first term counts the query and key projection weights and biases, and the second counts the channel-wise gates.
With $L=28$, $D=1152$, and $r=64$, the router adds $4.16$M and $4.23$M parameters for two and three stages, respectively, approximately $0.6\%$ of the $675$M parameters in SiT-XL/2.
The added computation is dominated by the width-$r$ projections; attention and gated aggregation operate only along the short stage axis at each patch position.

\section{Distillation Details}
\label{app:fist_distillation}

\paragraph{Teacher sampling.}
The frozen teacher is a SiT-XL/2 model pretrained with REPA \citep{yu2025repa}, whose checkpoint also provides the pretrained weights for FiST's backbone.
Each trajectory is generated with $N=20$ Heun integration steps on a uniform time grid, using classifier-free guidance \citep{ho2022cfg}.
Guidance takes the form $v_{\varnothing}+w(v_c-v_{\varnothing})$, where $v_c$ and $v_{\varnothing}$ denote conditional and unconditional velocity predictions.
Following the guidance settings used for REPA's best reported FID, the teacher sampler uses $w=1.8$ for $t\in[0,0.7]$ and $w=1$ elsewhere.

\paragraph{State selection.}
The selected states $\{\bar z_s\}_{s=0}^{K}$ correspond to $K+1$ evenly spaced times spanning the teacher's sampling interval $[0, 1]$.
When a selection time falls between two sampled times, its state is obtained by linear interpolation between the corresponding teacher states.
The initial noise $\bar z_0=\epsilon$ and final teacher endpoint $\bar z_K$ are retained exactly.

\paragraph{Loss reduction and optimization.}
Algorithm~\ref{alg:fist_distillation} summarizes distillation using batched operations.
Each MSE averages over the minibatch and latent coordinates; stage losses are summed before a single optimizer update.
For the augmented rollout, hidden features remain attached to the computation graph, allowing later losses to backpropagate through earlier stages.
The basic rollout omits the hidden-memory inputs, outputs, and updates.

\begin{algorithm}[tbp]
    \caption{Distillation-Based Initialization of FiST}
    \label{alg:fist_distillation}
    \begin{algorithmic}[1]
        \Require Stage count $K$, frozen teacher, initial FiST parameters $\Theta$
        \For{each distillation iteration}
            \State Sample a minibatch of noise--label pairs $(\epsilon,c)$.
            \State Generate teacher trajectories and select $\{\bar z_s\}_{s=0}^{K}$ as detached targets.
            \State Set $\mathcal{H}_{<1}=\varnothing$ and $\mathcal{L}=0$.
            \For{$s=1,\ldots,K$}
                \State $(\hat z_s,H_s)\gets F_{\Theta}(\bar z_{s-1},\mathcal{H}_{<s};e_s,c)$.
                \State $\mathcal{L}\gets\mathcal{L}+\operatorname{MSE}(\hat z_s,\bar z_s)$.
                \State $\mathcal{H}_{<s+1}\gets\mathcal{H}_{<s}\cup\{H_s\}$.
            \EndFor
            \State Update $\Theta$ using $\nabla_{\Theta}\mathcal{L}$.
        \EndFor
        \State \Return $\Theta$
    \end{algorithmic}
\end{algorithm}

\section{Refinement Details}
\label{app:fist_refinement}

\paragraph{Discriminator implementation.}
The frozen SiT-XL/2 backbone is evaluated through layer $16$, with conditioning dropout probability $0.1$, matching its pretraining setting.
Feature extraction uses each backbone's native clean-data time input:
$t=0$ for REPA-pretrained SiT and $t=1$ for vanilla SiT.
Let $\Phi_{\ell}(z,\tilde c)$ denote a feature map reshaped from the token sequence at selected layer $\ell\in\{4,8,12,16\}$, with $1152$ channels and spatial resolution $16\times16$.
Layer-specific $1\times1$ projections reduce these maps to $512$ channels after stacking:
\begin{equation}
    u_0=\sigma\!\left(
    \operatorname{GN}_0\!\left(
    \operatorname{Proj}_0\!\left(
    [\Phi_{\ell}(z,\tilde c)]_{\ell}
    \right)\right)\right),
    \label{eq:fist_discriminator_projection}
\end{equation}
where $[\cdot]_{\ell}$ denotes stacking along the selected-layer axis, $\sigma$ denotes SiLU, and GroupNorm is applied separately to each feature map throughout the trainable module.

Three residual blocks combine spatial processing within each map with mixing across the selected layers:
\begin{equation}
    \begin{aligned}
        \tilde u_b
        &=\sigma\!\left(
        \operatorname{GN}_{b,1}\bigl(
        \operatorname{Conv}_{b,1}(u_b)\bigr)\right),\\
        \Delta u_b
        &=\operatorname{Expand}_b\!\left(
        \operatorname{Mix}_b\!\left(
        \sigma\!\left(\operatorname{GN}_{b,\mathrm{mix}}\bigl(
        \operatorname{Squeeze}_b(\tilde u_b)\bigr)\right)\right)\right),\\
        u_{b+1}
        &=\sigma\!\left(
        \operatorname{Skip}_b(u_b)
        +\operatorname{GN}_{b,2}\bigl(
        \operatorname{Conv}_{b,2}(\tilde u_b+\Delta u_b)\bigr)\right),
        \quad b=0,1,2.
    \end{aligned}
    \label{eq:fist_discriminator_blocks}
\end{equation}
Here $\operatorname{Conv}_{b,1}$ and $\operatorname{Conv}_{b,2}$ are layer-specific $3\times3$ convolutions, implemented as grouped convolutions with one group per selected layer.
$\operatorname{Squeeze}_b$ and $\operatorname{Expand}_b$ are layer-specific $1\times1$ projections that reduce the channel width to one quarter and restore it, respectively.
$\operatorname{Mix}_b$ is a dense $1\times1$ projection over the joint layer--channel axis at each spatial position.
$\operatorname{Skip}_b$ is the identity when dimensions match; otherwise, it applies average pooling and a layer-specific $1\times1$ projection.
The three blocks have overall spatial strides $(1,2,2)$ and output channel widths $(512,768,1024)$, yielding spatial resolutions $(16\times16,8\times8,4\times4)$, respectively.

The layer axis is retained until final pooling.
A layer-specific $1\times1$ projection, GroupNorm, and SiLU precede averaging over the layer and spatial axes:
\begin{equation}
    \begin{aligned}
        f
        &=\operatorname{Mean}_{\ell,h,w}\!\left[
        \sigma\!\left(\operatorname{GN}_{\mathrm{out}}\bigl(
        \operatorname{Proj}_{\mathrm{out}}(u_3)\bigr)\right)
        \right],\\
        \bigl(d_{\omega}(z,\tilde c),q_{\omega}(z,\tilde c)\bigr)
        &=\bigl(
        \operatorname{Head}_{\mathrm{adv}}(f),
        \operatorname{Head}_{\mathrm{cls}}(f)
        \bigr).
    \end{aligned}
    \label{eq:fist_discriminator_readout}
\end{equation}
The two linear heads produce one adversarial logit and $1000$ class logits.
Spectral normalization \citep{miyato2018spectral} is applied to all trainable convolutional projections.

\paragraph{Loss reduction and weights.}
Each loss term is a batch mean; the discriminator adversarial loss sums the real and generated means.
For equally sized minibatches, near-zero adversarial logits, and near-uniform class probabilities over $C=1000$ classes, balancing initial gradient norms at the head outputs gives
\begin{equation}
    \lambda_{\mathrm{cls}}^{\mathrm{D}}
    \approx\frac{1}{\sqrt{2(1-1/C)}}\approx0.7,
    \qquad
    \lambda_{\mathrm{cls}}^{\mathrm{G}}
    \approx\frac{1}{2\sqrt{1-1/C}}\approx0.5.
    \label{eq:fist_classification_weight_calibration}
\end{equation}
At initialization, the per-sample adversarial and classification gradient norms at the head outputs are approximately $1/2$ and $\sqrt{1-1/C}$, respectively.
The discriminator includes separate real and generated adversarial logits, giving a combined gradient norm $\sqrt{2}$ times that of the generator.
The rounded classification weights remain fixed throughout training.

\paragraph{Alternating updates.}
Algorithm~\ref{alg:fist_refinement} implements the updates using Eqs.~\eqref{eq:fist_discriminator_loss} and~\eqref{eq:fist_generator_loss}.

\begin{algorithm}[tbp]
    \caption{End-to-End Refinement of FiST}
    \label{alg:fist_refinement}
    \begin{algorithmic}[1]
        \Require Distilled generator $G_{\Theta}$, frozen encoder $E$, frozen feature extractor $\Phi$, trainable discriminator module $\mathcal{D}_{\omega}$
        \For{each refinement iteration}
            \State Sample images $x$ with labels $c$, encode $z^{+}=E(x)$, and sample Gaussian noise $\epsilon$.
            \State Generate detached endpoints $z^{-}=\operatorname{sg}[G_{\Theta}(\epsilon,c)]$.
            \State Compute $\mathcal{L}_{\mathrm{D}}$ using Eq.~\eqref{eq:fist_discriminator_loss} and update $\omega$ using $\nabla_{\omega}\mathcal{L}_{\mathrm{D}}$; hold $\Theta$ fixed.
            \State Sample new Gaussian noise $\epsilon$, retaining the labels $c$.
            \State Generate endpoints $z^{-}=G_{\Theta}(\epsilon,c)$ with gradient tracking enabled.
            \State Compute $\mathcal{L}_{\mathrm{G}}$ using Eq.~\eqref{eq:fist_generator_loss} and update $\Theta$ using $\nabla_{\Theta}\mathcal{L}_{\mathrm{G}}$; hold the updated $\omega$ fixed.
        \EndFor
        \State \Return $\Theta$
    \end{algorithmic}
\end{algorithm}

\paragraph{Gradient propagation with hidden memory.}
For the augmented rollout, combine the latent state and hidden memory into
\begin{equation}
    \xi_s=(z_s,\mathcal{H}_{<s+1}),\qquad
    \xi_s=\widetilde F_{\Theta}^{s}(\xi_{s-1};c),\qquad
    \xi_0=(\epsilon,\varnothing),
    \label{eq:fist_augmented_state}
\end{equation}
where $\widetilde F_{\Theta}^{s}$ applies $F_{\Theta}$ at stage $s$ and appends the returned hidden features to memory.
For one sampled endpoint loss, let $v_s=\nabla_{\xi_s}\ell_{\mathrm{G}}$.
The gradient recursion becomes
\begin{equation}
    \begin{aligned}
        v_K&=(\nabla_{z_K}\ell_{\mathrm{G}},0),\\
        v_{s-1}
        &=\left(
        \frac{\partial \widetilde F_{\Theta}^{s}(\xi_{s-1};c)}
             {\partial \xi_{s-1}}
        \right)^{\!\mathsf T}v_s,
        \qquad s=K,\ldots,1,\\
        \nabla_{\Theta}\ell_{\mathrm{G}}
        &=\sum_{s=1}^{K}
        \left(
        \frac{\partial \widetilde F_{\Theta}^{s}(\xi_{s-1};c)}
             {\partial\Theta}
        \right)^{\!\mathsf T}v_s.
    \end{aligned}
    \label{eq:fist_augmented_gradient}
\end{equation}
The parameter derivatives hold the stage input $\xi_{s-1}$ fixed.
The terminal memory gradient is zero because the loss acts directly only on $z_K$; earlier hidden features receive gradients through their influence on subsequent stages.
This accounts for gradient propagation through both latent-state transitions and hidden features passed through the cross-stage router.
Omitting the memory recovers Eq.~\eqref{eq:fist_basic_shared_gradient}.

\section{Experimental Setup Details}
\label{app:experimental_setup}

\paragraph{Pretrained checkpoints.}
We use the EMA version (\texttt{sd-vae-ft-ema}) of the publicly available SD-VAE.
The REPA checkpoint is the officially released SiT-XL/2 model trained for 4M iterations \citep{yu2025repa}.
The vanilla SiT-XL/2 checkpoint used in the pretrained-checkpoint ablations comes from the official SiT release \citep{ma2024sit}.

\paragraph{Training configuration.}
Table~\ref{tab:fist_training_settings} summarizes the training hyperparameters.
Warmup and cosine decay are applied separately in the two training phases.
The minimum learning rate is $0.1$ times the corresponding base learning rate.
Settings varied in individual ablations are specified alongside their results.
Teacher sampling and state selection are described in Appendix~\ref{app:fist_distillation}, while discriminator implementation and loss-weight calibration are detailed in Appendix~\ref{app:fist_refinement}.

\begin{table}[t]
\centering
\caption{\textbf{Training hyperparameters.}
Each refinement iteration consists of one discriminator update followed by one generator update.
G and D denote the generator and discriminator, respectively.}
\label{tab:fist_training_settings}
\begingroup
\small
\setlength{\tabcolsep}{5pt}
\renewcommand{\arraystretch}{1.08}
\begin{tabular*}{\linewidth}{@{\extracolsep{\fill}}l c c@{}}
\toprule
Setting & Distillation & Refinement \\
\midrule
Training duration & 40K steps & 20K iterations \\
Batch size & 1024 & 1024 \\
Learning rate & $10^{-4}$ & $10^{-5}$ (G), $4\times10^{-5}$ (D) \\
Optimizer & AdamW & AdamW \\
AdamW betas & $(0.9,0.99)$ & $(0.9,0.99)$ \\
Weight decay & 0 & 0 \\
Learning-rate schedule & Cosine decay & Cosine decay \\
Warmup steps & 1K & 1K \\
Minimum learning-rate ratio & 0.1 & 0.1 \\
Maximum gradient norm & 1.0 & 1.0 \\
\bottomrule
\end{tabular*}
\endgroup
\end{table}

\clearpage
\section{Additional Qualitative Samples}
\label{app:additional_samples}

We present uncurated samples from 20 selected ImageNet classes.
All samples are generated by three-stage augmented FiST at the
checkpoint with FID 1.11 and IS 282.0.

\showcase{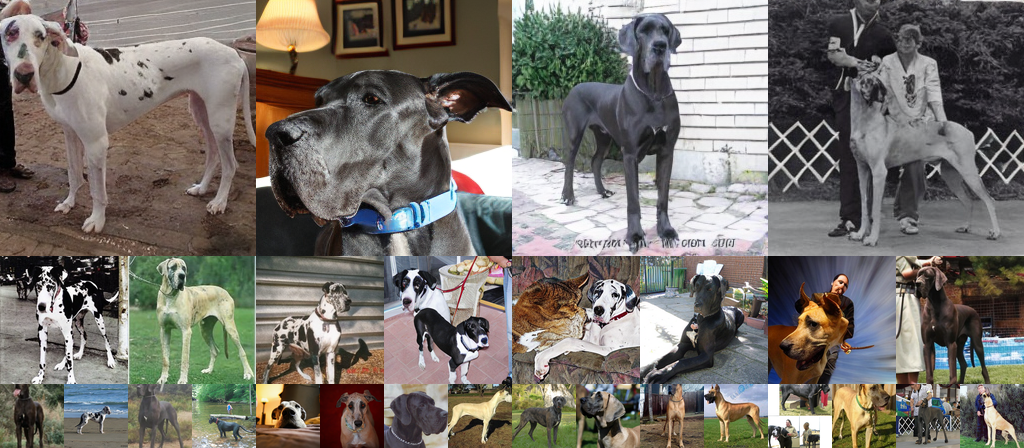}{Great Dane}
\showcase{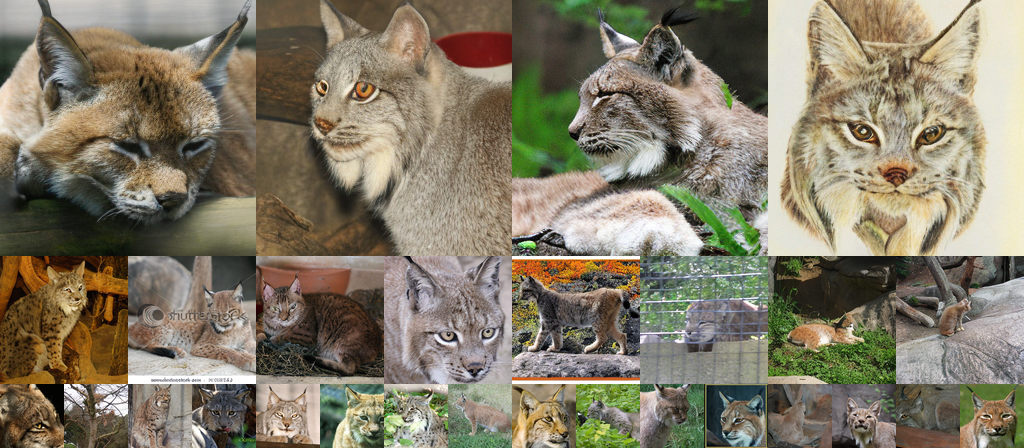}{lynx}
\showcase{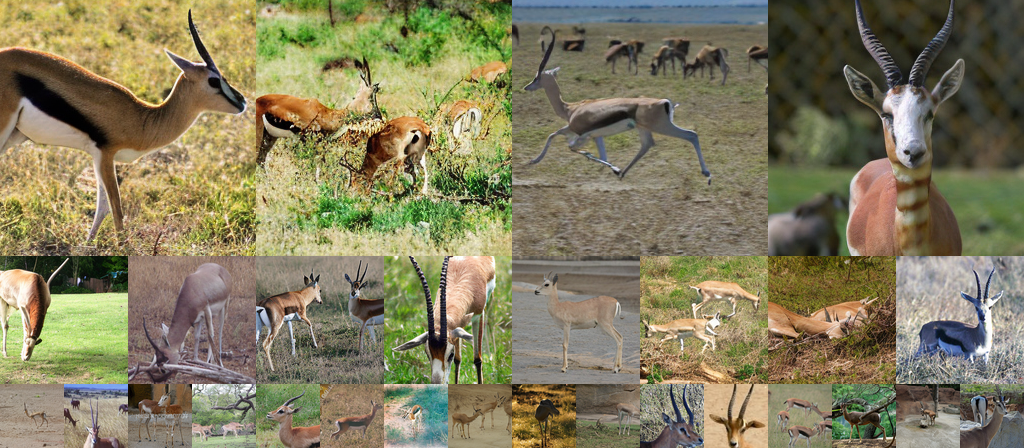}{gazelle}
\showcase{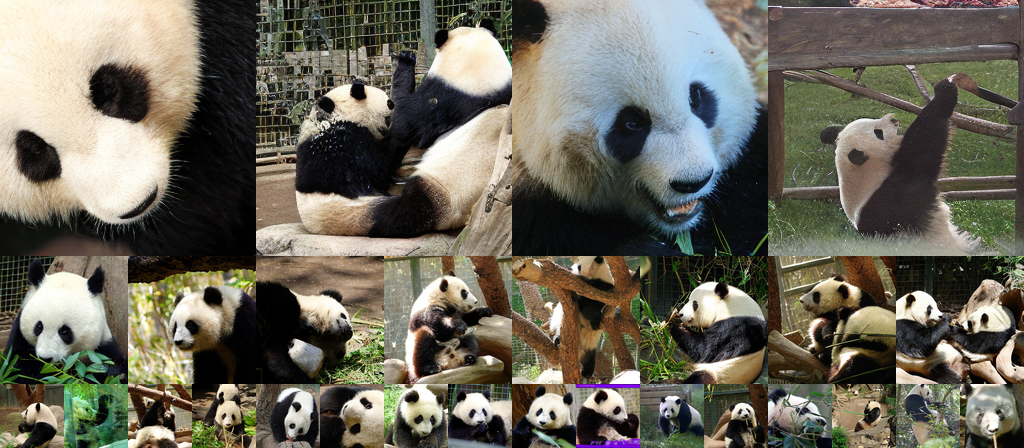}{giant panda}
\showcase{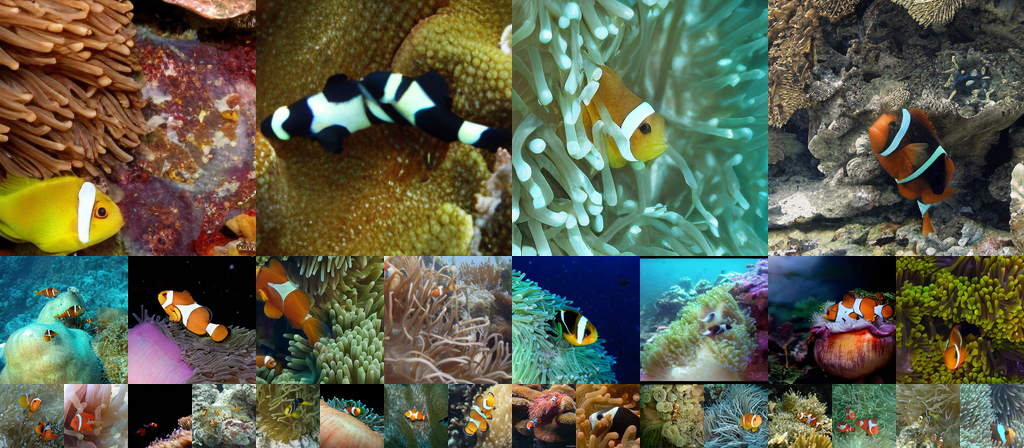}{anemone fish}
\showcase{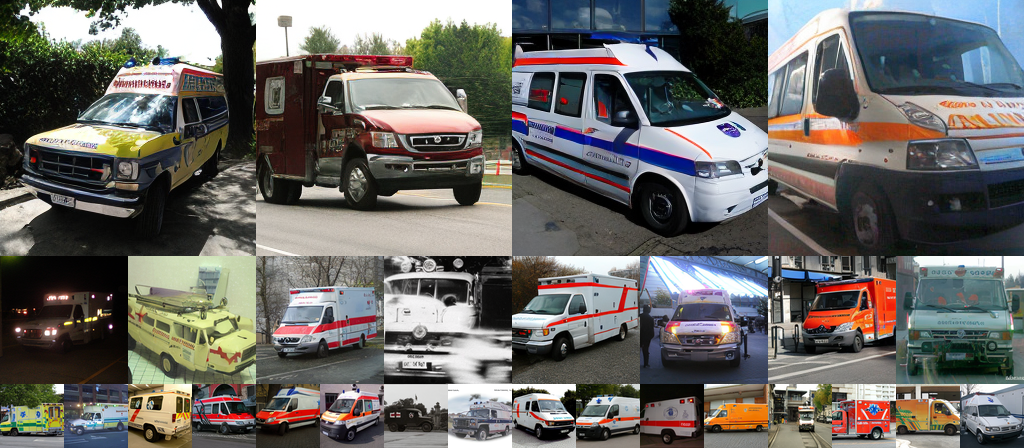}{ambulance}
\showcase{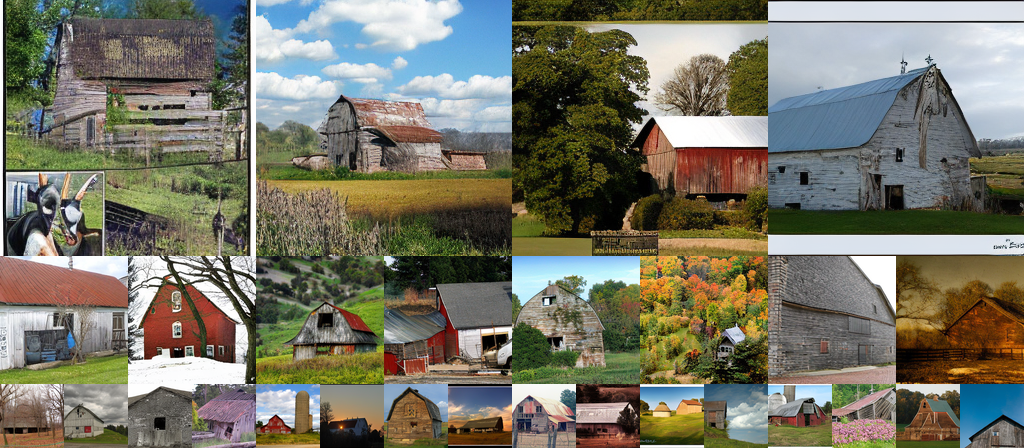}{barn}
\showcase{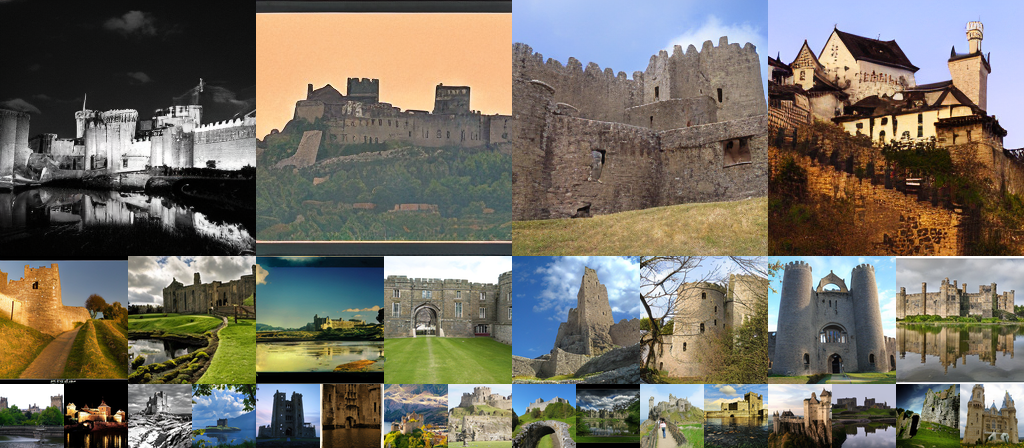}{castle}
\showcase{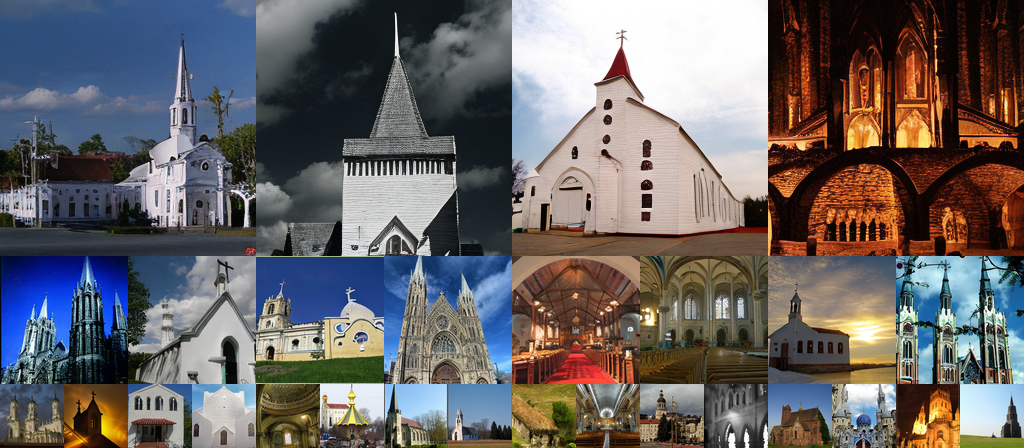}{church}
\showcase{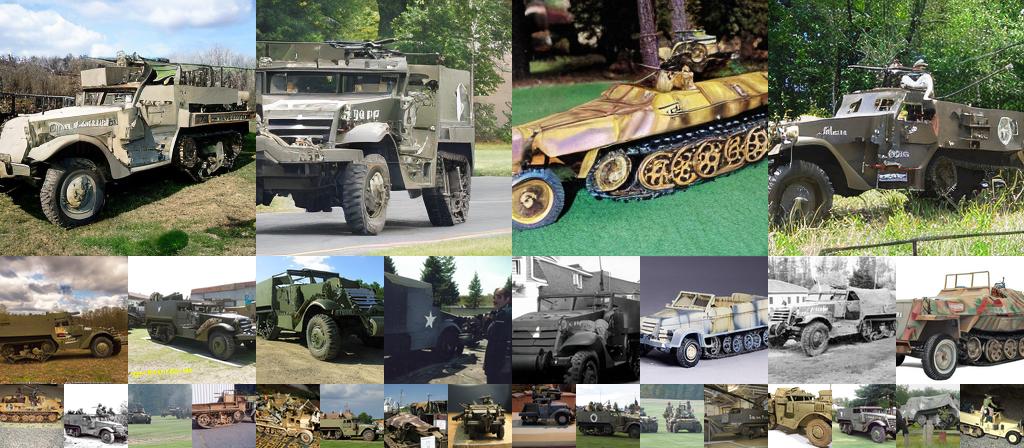}{half track}
\showcase{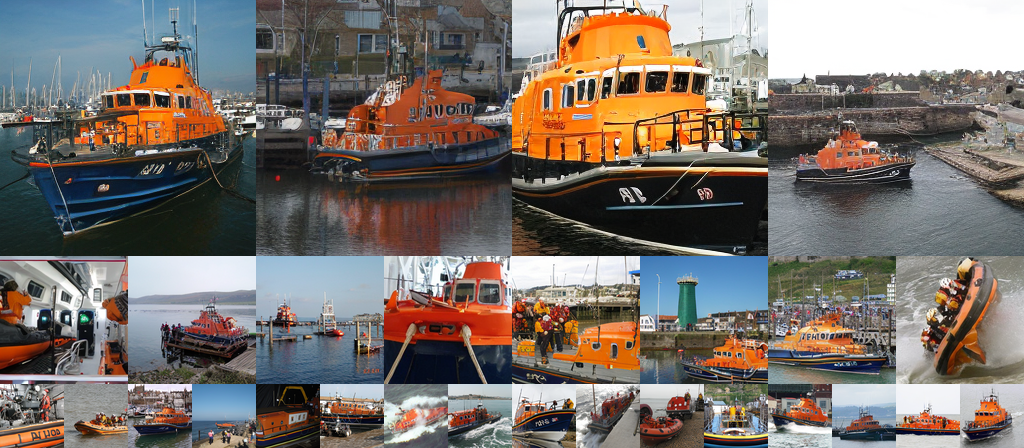}{lifeboat}
\showcase{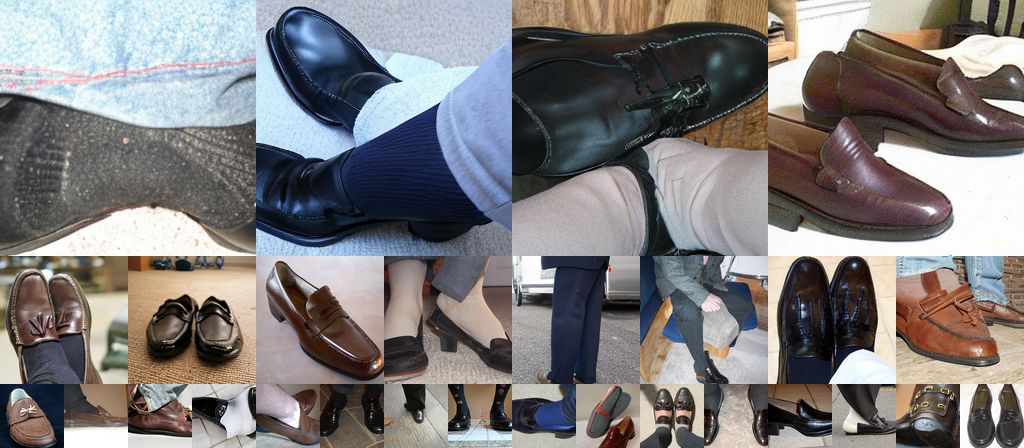}{Loafer}
\showcase{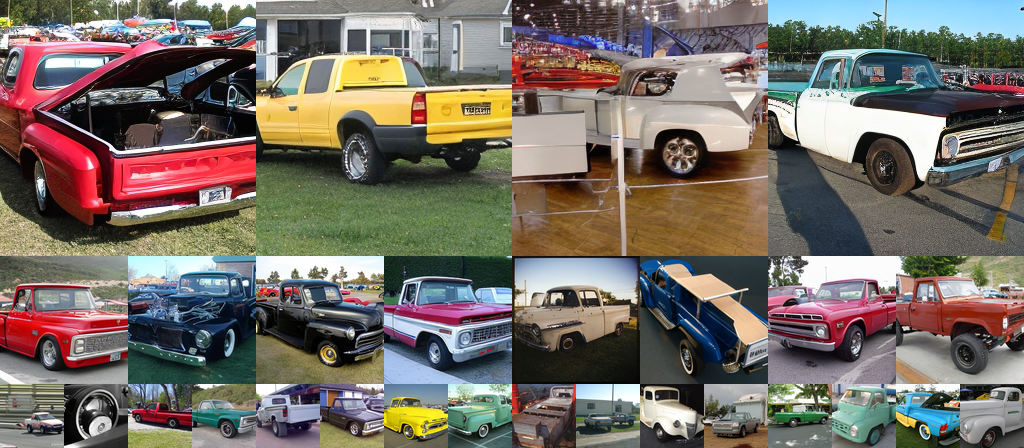}{pickup}
\showcase{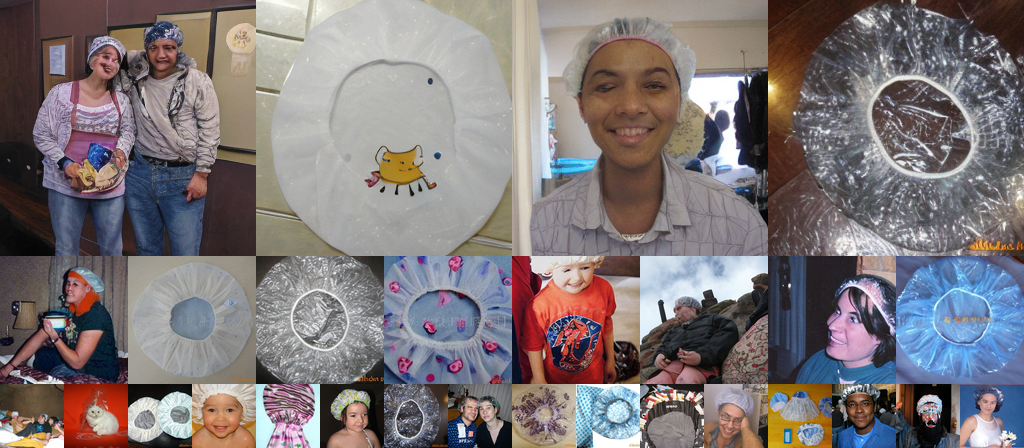}{shower cap}
\showcase{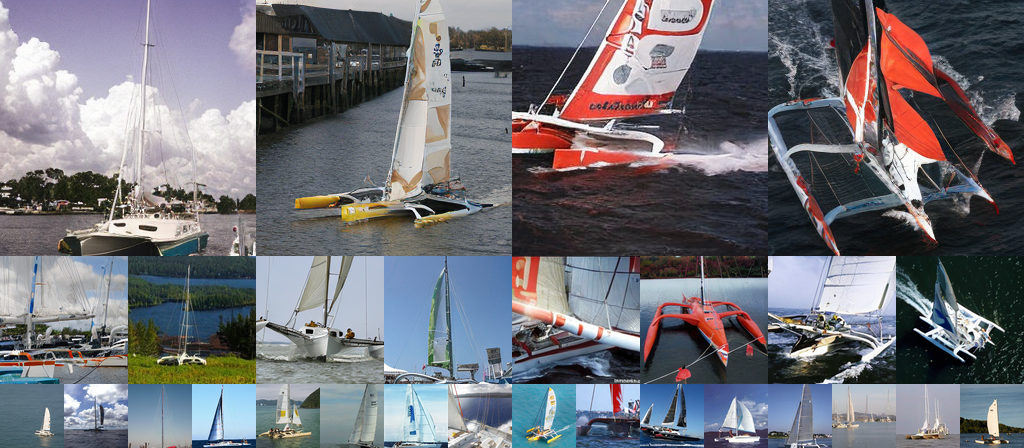}{trimaran}
\showcase{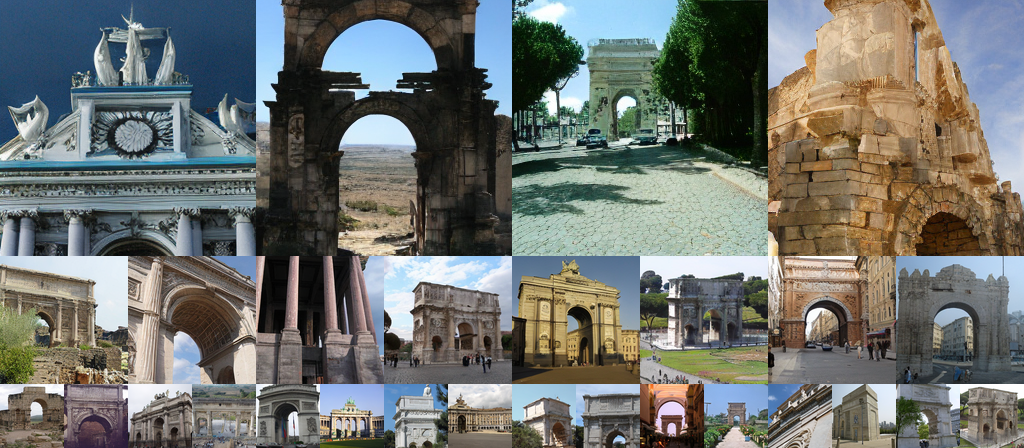}{triumphal arch}
\showcase{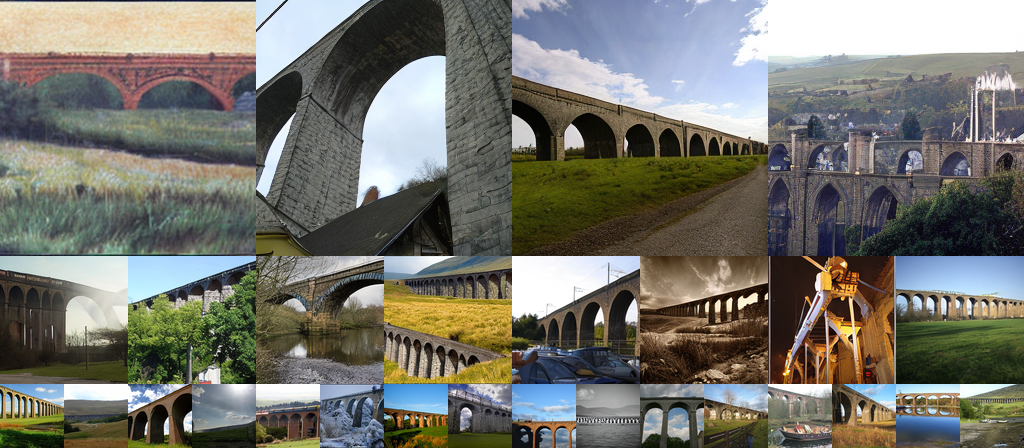}{viaduct}
\showcase{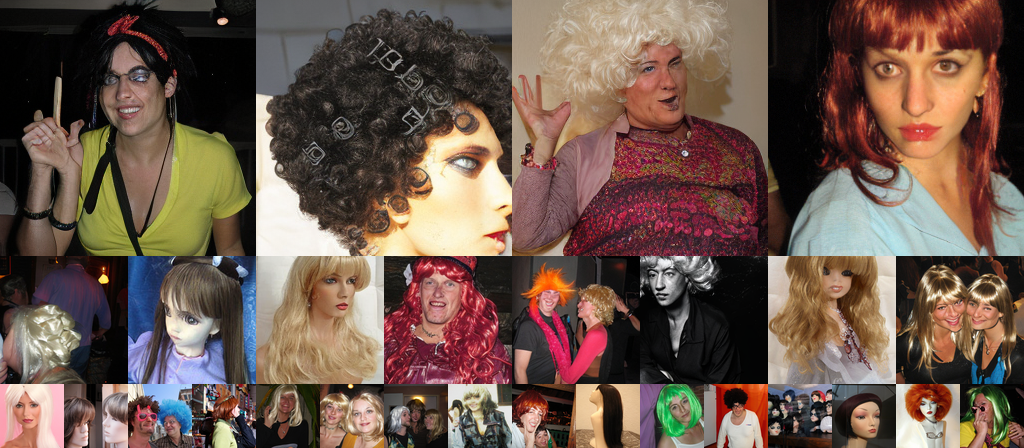}{wig}
\showcase{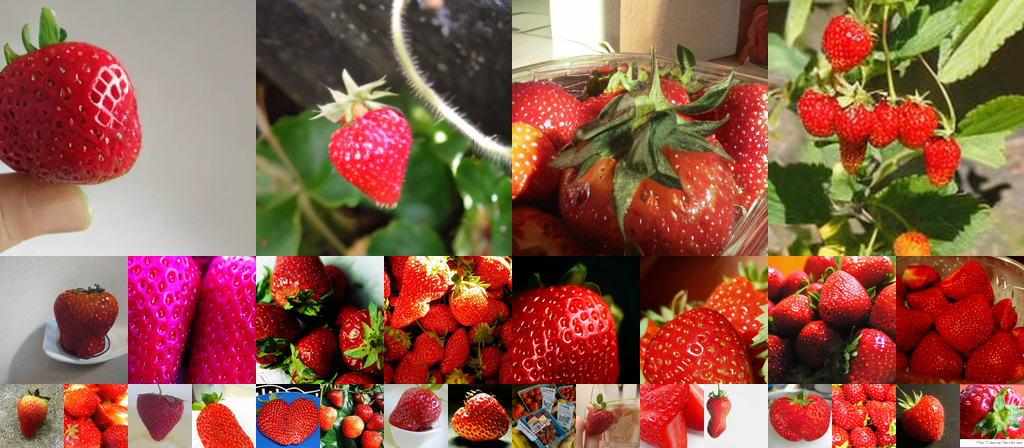}{strawberry}
\showcase{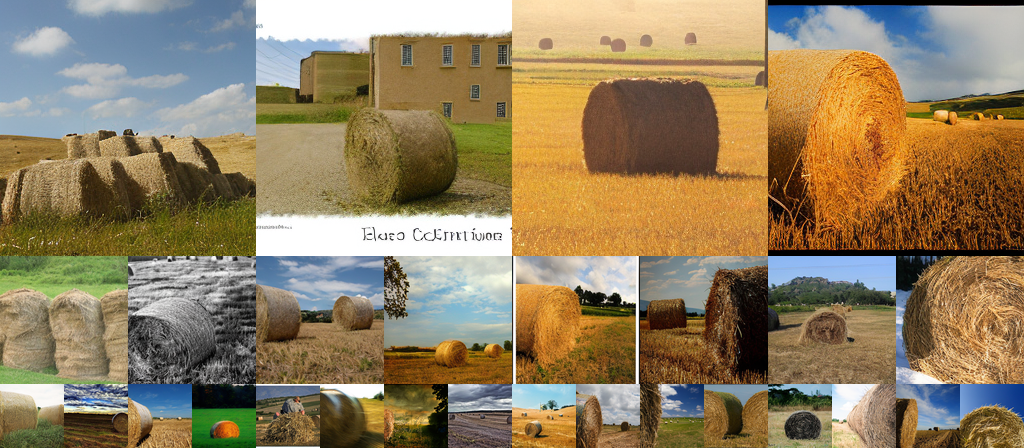}{hay}

\end{document}